\documentclass[letterpaper]{article} 
\usepackage{aaai25}  
\usepackage{times}  
\usepackage{helvet}  
\usepackage{courier}  
\usepackage[hyphens]{url}  
\usepackage{graphicx} 
\usepackage{natbib}  
\usepackage{caption} 
\usepackage{algorithm}
\usepackage{algorithmic}

\usepackage{newfloat}
\usepackage{listings}
\DeclareCaptionStyle{ruled}{labelfont=normalfont,labelsep=colon,strut=off} 
\floatstyle{ruled}
\newfloat{listing}{tb}{lst}{}
\floatname{listing}{Listing}
\usepackage{amsmath}
\usepackage{amssymb}
\usepackage[toc,page]{appendix}
\usepackage{tikz}
\usepackage{amsthm}
\usepackage{booktabs}
\usepackage{pdfpages}

 \nocopyright

\title{Harnessing Event Sensory Data for Error Pattern Prediction in Vehicles:\ A Language Model Approach}
\author{
   Hugo Math, Rainer Lienhart, Robin Schön \\
}
\affiliations{
    Augsburg University, Augsburg 86159, Germany
}

\usepackage{bibentry}

\begin{document}

\maketitle

\begin{abstract}
In this paper, we draw an analogy between processing natural languages and processing multivariate event streams from vehicles in order to predict \textit{when} and \textit{what} error pattern is most likely to occur in the future for a given car. Our approach leverages the temporal dynamics and contextual relationships of our event data from a fleet of cars. Event data is composed of discrete values of error codes as well as continuous values such as time and mileage. Modelled by two causal Transformers, we can anticipate vehicle failures and malfunctions before they happen. Thus, we introduce \textit{CarFormer}, a Transformer model trained via a new self-supervised learning strategy, and \textit{EPredictor}, an autoregressive Transformer decoder model capable of predicting \textit{when} and \textit{what} error pattern will most likely occur after some error code apparition.
Despite the challenges of high cardinality of event types, their unbalanced frequency of appearance and limited labelled data, our experimental results demonstrate the excellent predictive ability of our novel model. Specifically, with sequences of $160$ error codes on average, our model is able with only half of the error codes to achieve $80$\% F1 score for predicting \textit{what} error pattern will occur and achieves an average absolute error of $58.4 \pm 13.2$h \textit{when} forecasting the time of occurrence, thus enabling confident predictive maintenance and enhancing vehicle safety.
\end{abstract}

%
 \begin{links}
     \link{Code}{https://github.com/Mathugo/AAAI2025-CarFormer-EPredictor}
 \end{links}

\section{Introduction}
Today's vehicles generate an astounding amount of data, typically reported as events on an irregular basis, but continuously over time.
Some events occur simultaneously, while others are scattered over time, with unequally distributed time intervals between their occurrence. They usually report numerical and/or categorical features. 
In our work, we focus on processing and analyzing such \textit{multivariate and irregular event streams} produced by modern cars, for which related research is still sparse. Our sequences of discrete events in time are 
known as DTCs (\textit{diagnostic trouble codes}). DTCs are preferred over raw sensory data because they provide less granular and discrete information. This makes them easier to analyze. 

\begin{figure}[t]
\centering
\includegraphics[width=1\columnwidth]{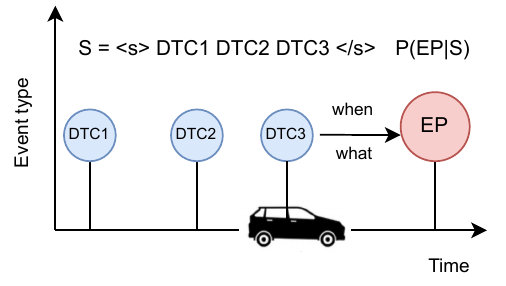} 
\caption{Error pattern (EP) prediction (when and what with which probability) based on the past sequence $S$ of diagnostic trouble codes (DTCs).}
\label{car}
\end{figure}

Our goal is to learn the correlation between DTCs using a DTC-based language model to predict \textit{when} and \textit{what} with which \textit{probability} an error pattern (EP) is occurring after having seen a number of DTCs (see Figure \ref{car}). Thus, we consider DTCs to be the words in our language. In contrast, EPs are very different from DTCs. They are defined by domain experts after observing DTC sequences. Therefore, they are way more precise about the critical error that the car is having. While some DTCs can also be noisy and repetitive events about recurring errors  (e.g. electrical issues, software updates), EPs characterize a whole error sequence consisting of precise vehicle failures (e.g. engine or battery failures).
Recent research has approached predictive maintenance in the automotive field using sequences of DTCs via RNNs 
\cite{faultpredmultivariatevehicule} and, more recently, Transformers \cite{Hafeez2024DTCTranGruIT} to predict the next DTC. However, distinguishing minor errors or noise-like DTCs from important events such as error patterns (EPs) is essential since the latter poses higher risks and necessitates greater safety and maintenance measures, such as vehicle immobilization or addressing critical malfunctions.
Furthermore, as the data volume increases, accurately predicting the next DTC becomes challenging, particularly when the event type cardinality approaches $\sim 10^4$. This phenomenon is akin to language processing, where the accuracy of the next token prediction using greedy decoding or other methods exponentially decreases with sequence length and vocabulary size due to error accumulation \cite{bachmann2024pitfallsnexttokenpred}.

Historically, Hawkes Processes and their neural variants have advanced the state of the art in event modelling for next event and time prediction tasks \cite{hawkeppp, rrnembedding, nppreview}. Transformer-based models like BERT \cite{bert} and GPT-3 \cite{gpt3} have gained overwhelming popularity due to their attention-based architecture, flexibility, parallelization, and state-of-the-art performance in sequence modelling. Consequently, models adapted to discrete-time sequences using Transformers have emerged naturally \cite{selfatthawke, transformerhawkeprocess, shou2024selfsupervisedcontrastivepretrainingmultivariate}, achieving state-of-the-art performance in next event prediction benchmarks.
By leveraging the sequential nature of our data, we can define sentences as the concatenation of each discrete event:
$$ "<s> \ DTC1 \ DTC2 \ .. \ DTCn \ </s>" $$\label{DTCCONCAT}
as shown in Figure \ref{car}, and embed it into $\mathbb{R}^D$. We make several modifications to the vanilla Transformer from \cite{attention}, incorporating continuous-time and mileage positional embeddings as additional context. Using two distinct training phases, we introduce CarFormer a pre-trained model acting as an encoder and EPredictor a decoder Transformer-based model that generates a probability distribution over a set of error patterns for each event step \(i\) to determine \textit{what} EPs will most likely happen and estimates a time for \textit{when} it will occur.
\subsection{Contributions}
To the best of our knowledge, introducing a causal Transformer model for error pattern and DTC prediction (event type and time) has not been explored, despite some related work on event-data and next-DTC prediction \cite{shou2024selfsupervisedcontrastivepretrainingmultivariate, Hafeez2024DTCTranGruIT, faultpredmultivariatevehicule}. The main contributions of this paper are:
\begin{itemize}
\item \textbf{CarFormer}: An encoder Transformer-based model designed to ingest scattered continuous event streams from vehicles, trained via a multi-task learning strategy. This model will transform the DTC-sequences into hidden representations that can be processed by the EPredictor.
\item \textbf{EPredictor}: An autoregressive decoder Transformer-based model that specializes in predictive maintenance, particularly error patterns by estimating  \textit{when} and \textit{what} error patterns will most likely occur.
\end{itemize}

\section{Background and Related Work}
\subsection{Event Sequence Modelling with TPP}
Event data is commonly modelled via Temporal Point Processes (TPPs), which describe stochastic processes of discrete events. Each event is composed of a time of occurrence $t \in \mathbb{R}^+$ and an event type $u \in U$ forming a pair $(t, u)$. $U$ is a finite set of discrete event types.
A sequence is constructed with multiple pairs of events, such as $S = \{(t_1, u_1), ... , (t_L, u_L)\}$ where $0<t_1< ... < t_n$.

TPPs are usually represented as a counting process $N(t) \ \forall t \geq 0$ for the events of type $u$, which describes the number of occurrences of an event over time. 
The goal is to predict the next event $(u', t')$ given the history $H_t := \{(t_i, u_i) \in \mathbb{R}^+ \times U|t_i<t\}$ of all events that occurred up to time $t$.
$ $
We define $\lambda^*$ to model the instantaneous rate of an event in continuous time. 
Thus, the probability of occurrence for an event $(u', t')$ is conditioned on the history of events $H_t$:
\begin{equation}
\begin{split}
   \lambda^* (t) d t &:= P((u', t'): t' \in [t, t+dt)|H_t) \\ & = \mathbb{E}(dN(t)|H_t) 
\end{split}
  \label{eq:intensity}
\end{equation}
which we could translate as the expected number of events during an infinitesimal time window $[t, t+d t) $ knowing the history $H_t$. We assume that two events do not occur simultaneously, i.e., $d N(t) \in \{0, 1\}$.

The Hawkes Process \cite{hawkeppp} has been arguably the most studied modelling technique for TPPs. It assumes some parametric form of the conditional intensity function $\lambda^*(t)$ and states that an event excites future events additively and decays using a function $f$ over time. For example:
\begin{equation}
\lambda^*(t) = \mu+\sum_{(u_i, t_i) \in H_t} f(t-t_i) 
\end{equation}
where $\mu \geq 0$ is the base intensity.
Neural TPPs, on the other hand, aim at reducing the inductive bias of the Hawkes Process, which states that past events excite future ones additively. They do this by approximating $\lambda^*$ using a neural network (RNN, LSTM, Transformer) \cite{selfatthawke, rrnembedding}. 
Recent papers suggest using a generative and contrastive approach for event sequence modeling, showing promising results across predictive benchmarks. \cite{lin2022exploring} uses next-event prediction as their main training objective, while \cite{shou2024selfsupervisedcontrastivepretrainingmultivariate} have three specific objectives plus a contrastive loss.
\subsection{Transformers for Event Streams}
Encoding the history $H_t$ into historical hidden vectors using a Transformer enhanced the performance on event prediction benchmarks as shown in  \cite{transformerhawkeprocess} with the Transformer Hawkes Process (THP) or the self-attentive Hawkes Process \cite{selfatthawke}. They typically reused the vanilla Transformer \cite{attention} and created two embeddings: 

\subsubsection{Time Embedding} Time embedding replaces the traditional positional encoding which grants the Transformer model positional information of each token within the sequence.
This time embedding is defined deterministically with periodic functions exactly like in \cite{attention}:

\begin{equation}
\mathbf{P}_{i,j} := \begin{cases}
    \sin(t_i \times \omega_0^{j/d}) & \text{if } j\mod 2 = 0 \\
    \cos(t_i \times \omega_0^{(j-1)/d}) & \text{if } j\mod 2 = 1\\
\end{cases}
\end{equation}
where $i$ is the index of the $i$-th event, $\omega_0$ is the frequency (usually $10^{-4}$).
\subsubsection{Event-type Embedding}\label{eventypeemb} To get a dense representation of our sequence, we embed each event into a $d$ dimensional space using an embedding matrix $\mathbf{L}^{V \times d}$ where $V$ is the distinct number of events (vocabulary). As we would do for word embeddings, we create a sequence of one-hot encoded vectors from the event types $\{u_i\}^L_{i=0}$ as $\mathbf{Y} \in \mathbb{R}^{L \times V}$. Thus, the event-type embedding $\mathbf{E} = \mathbf{Y} \mathbf{L} \in \mathbb{R}^{L \times d}$ and the input embedding $\mathbf{U}$ is defined as $\mathbf{U} =\mathbf{E} + \mathbf{P} \in \mathbb{R}^{L \times d}$

\subsubsection{Attention}
The majority of research utilizing Transformer models for sequence data employs the architecture introduced by \cite{attention}.
We define three linear projection matrices $\mathbf{Q} =\mathbf{U} \mathbf{W}^Q$, $\mathbf{K} =\mathbf{U} \mathbf{W}^K$, and $\mathbf{V} =\mathbf{U} \mathbf{W}^V$. They are called \textit{query}, \textit{key}, and \textit{value}, respectively.
$\mathbf{W}^Q$, $\mathbf{W}^K$, $\mathbf{W}^V$ are trainable weights. Essentially $\mathbf{Q}$ represents what the model is looking based on the input $\mathbf{U}$, $\mathbf{K}$ is the label for the input's information and $\mathbf{V}$ is the desired representation of the input's semantics. 
The attention score can be computed as:
\begin{equation}
\label{eq:vanillattention}
 \mathbf{A} = \text{softmax}(\mathbf{Q}\mathbf{K}^T/\sqrt{d}) 
\end{equation}
\begin{equation}
\mathbf{C} = \mathbf{A} \mathbf{V}
\end{equation}
where $d$ denotes the number of attention heads and $\mathbf{A} \in \mathbb{R}^{L \times L}$ the attention scores of each event pair $i,j$. A final hidden representation $H$ is obtained via a layer normalization (LayerNorm), a pointwise feed-forward neural network (FFN) and residual connections via: 
\begin{equation}
\begin{split}
    \mathbf{U}'= \text{LayerNorm}(\mathbf{C}+\mathbf{U}) \\
    H = \text{LayerNorm}(\mathbf{U'}+\text{FFN}(\mathbf{U}))
\end{split}
\end{equation}
\subsection{Self-supervised learning}
By leveraging an efficient pre-training task (e.g. token masking or next token prediction) and then fine-tuning a smaller model with fewer parameters for specific tasks, Transformer-based models achieve state-of-the-art performance in natural language processing tasks \cite{bert, gpt3}.
For example, the GPT model is pre-trained on the next token prediction task, where the labels are generated by shifting the tokens to the right. Then, a classification head is added on top of the Transformer model to assign a probability to each token. Contrary to BERT \cite{bert} a causal mask is applied so that tokens can only attend to the previous one, thus preventing cheating. In section \ref{sec:pretraining} we will introduce a pre-trained model serving as an encoder trained on specific event prediction tasks to ensure adaptability to the event stream domain of vehicles. We found that modelling event streams with autoregressive Transformer-based models for fault predictions were not addressed well in the literature \cite{nppreview} although there is some related work for TPPs \cite{lin2022exploring, shou2024selfsupervisedcontrastivepretrainingmultivariate}. 

\section{Data}

\begin{table}[ht]
    \centering
    \begin{tabular}{|c|p{6cm}|}
        \hline
        \textbf{Notation} & \textbf{Description} \\
        \hline
        $u$ & Discrete event type, in our case a DTC. \\
        \hline
        $d$ & Absolute mileage of the vehicle in km.\\
        \hline
        $m$ & Mileage of the vehicle in km since the first DTC ($u_0$) occurred in a sequence $S$ such as $m_i = d_i - d_0$\\
        \hline
        $ts$ & Unix timestamp attached to each DTC.\\
        \hline
        $t$ & Number of hours passed since the first DTC occurred in a sequence. More specifically, \(t_i\) is defined as \(t_i = ts_i - ts_0\). \\
        \hline
        $S$ & Sequence of triplets (event type, time, mileage) defined as \(S = \{(u_i, t_i, m_i)\}_{i=0}^{L}\) of length $L$ with index starting from 0. \\
        \hline
        $i\in \{0,..., L\}$ & index of element in (event) sequence. \\
        \hline
    \end{tabular}
    \caption{List of symbols and their respective meanings}
    \label{tab:notation}
\end{table}

\subsubsection{Overview}
Diagnostic data is generated by various \textit{Electronic Control Units} (ECU) in a vehicle at irregular intervals. Diagnostic data differs substantially from raw sensor data, since diagnostic data or fault events are categorical and relate to various problems within the vehicle. We construct a \textit{Diagnosis Trouble Code} (DTC) indicating the precise error from 3 pieces of information arriving at the same timestamp $ts$ and mileage $d$: (1) the ID number of the ECU, (2) an error code (\textit{Base-DTC}) and (3) a \textit{Fault-Byte}. A single DTC token is composed of these 3 elements:
$$ DTC = ECU|\textit{Base-DTC}|\textit{Fault-Byte} $$
Thus, we can encode uniquely each DTC by a single token to predict the next token directly.

This research uses an anonymized vehicular DTC sequence dataset of $1.7 \times 10^6$ sequences with on average 150 DTCs per sequence. Each sequence belongs to a unique vehicle.
In a sequence S, each DTC (commonly referred to as event type $u_i$) is attached with a time $t_i$ and a mileage $m_i$ constructing a single event $(u_i, t_i, m_i)$. You can find an overview of the DTC elements in Table \ref{tab:data_details_dtc}. 
\begin{table}[h!]
\centering
 \begin{tabular}{c c c} 
 \hline
 Data & \# of values & Description \\ [0.5ex] 
 \hline
 DTC & 8710 & Diagnostic Trouble Code  \\ 
 ECU & 61 & Electronic Control Unit  \\
 Base-DTC & 7726 & Error Code \\
 Fault-Byte & 2 & Binary Value  \\
 \hline
 \end{tabular}
 \caption{Number of Distinct values of the DTC elements}
 \label{tab:data_details_dtc}
\end{table}

To get a full sequence $S = \{(u_i, t_i, m_i)\}^L_{i=0}$, we obtain the last known timestamp $ts_L$ and mileage $d_L$ and select all DTCs that are no further than: (1) a given period in the past ($ts_L-ts_i \leq 30$ days) and (2) a given distance in the past ($d_L-d_i \leq 300$km). Table \ref{tab:notation} explains the different data notation.

\subsubsection{Time and Mileage}
\begin{figure}[t]
    \centering
    \includegraphics[width=1\columnwidth]{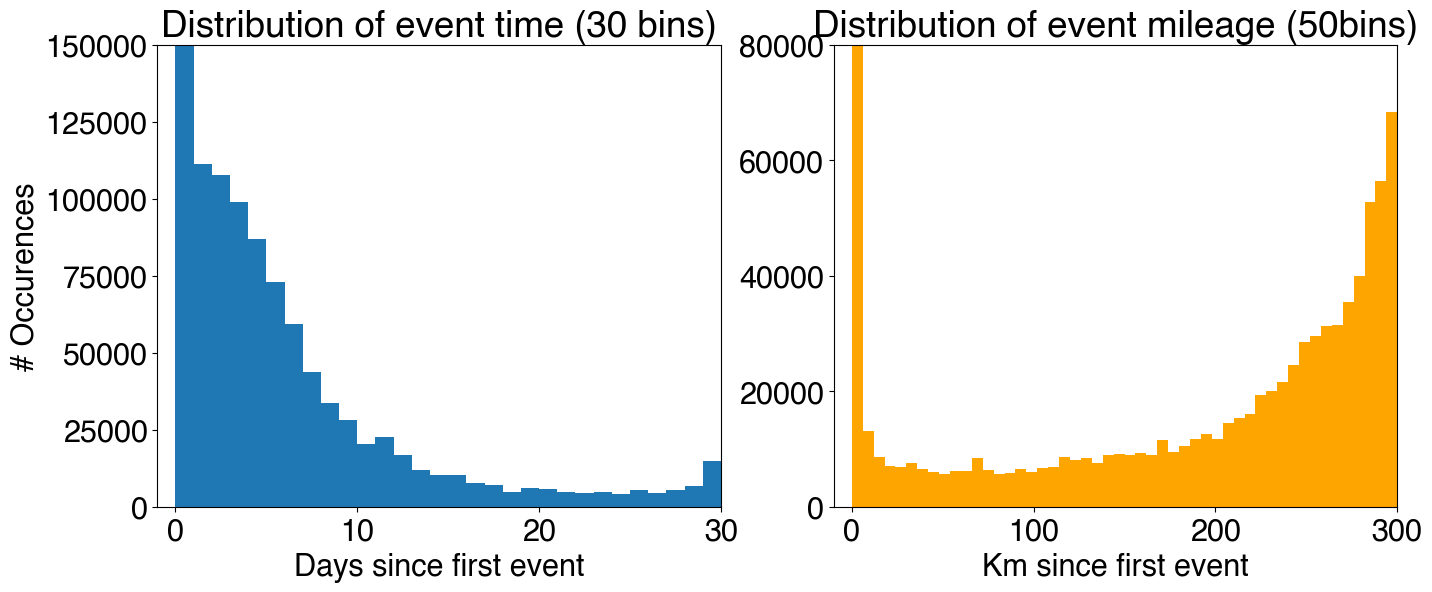}
    \caption{Distribution of $t_i$ and $m_i$ in our data set} 
    \label{fig:aai_time_mileage}
\end{figure}

We draw the distribution of $t_i$ and $m_i$ in Figure \ref{fig:aai_time_mileage}.
We observe peaks at 0 on both distributions due to truncation and missing values. 
In order to feed our model with the time $t$ feature, we need a scaling method to level out the left-tail distribution somewhat. Using $\log_b (t + 1)$ is a natural choice. At the same time, we want to approximately map $t$ into the range of $[-1, 1]$. Therefore, we apply the following non-linear function $f_t: \mathbb{R}^+ \to \mathbb{R}$ to $t$: 
\begin{equation}\label{eq:ttransform}
    t' = f_t(t, b) = \log_b (t + 1) - 1\: \forall t \in \mathbb{R}^+
\end{equation}
with $b$ chosen appropriately and feed  $t'$ into a neural network. 
In TPPs the time is usually represented as inter-event time or its logarithm \cite{nppreview, rrnembedding}.

\section{Pre-training}\label{sec:pretraining}
\subsection{Embeddings}\label{section:embeddings}
CarFormer uses 4 different embeddings to capture spatial and temporal dependencies of irregular event apparition. These embeddings differ from the positional embedding $\mathbf{P}$ used in TPPs  \shortcite{transformerhawkeprocess}, \shortcite{multilabelpredictionfault} and next-DTC prediction studies \shortcite{Hafeez2024DTCTranGruIT}, \shortcite{faultpredmultivariatevehicule}. We embed both time $t$ and mileage $m$ and use a rotation matrix to induce absolute and relative event positions for our Transformer such as: 
\begin{itemize}
  \item \textbf{Event-type embedding} $\mathbf{E} \in \mathbb{R}^{L \times d}$ is obtained like described in section \ref{eventypeemb}.
  \item \textbf{Absolute time embedding} $\mathbf{T} \in \mathbb{R}^{L \times d}$ is constructed on-the-fly at each forward pass by a linear transformation $t_{i,j} = t'_i w_j + b_j$ where $w_j, b_j$ are learnable parameters and $t'_i$ is the scaled time at event step $i$.
  \item \textbf{Mileage embedding} $\mathbf{M} \in \mathbb{R}^{L \times d}$ is obtained via a learnable lookup table $\mathbf{W}^{m_{\text{max}} \times d}$ where $m_\text{max}=300$km. Each row $\mathbf{w}_m \in \mathbb{R}^{d}$ corresponds to the learnable embedding vector for the discrete mileage $m$. The continuous mileage $m_i \in \mathbb{R}^{+}$ is cast to an integer value $m=\lfloor m_i \rfloor$. 
    
  \item \textbf{Rotary Position Embedding (RoPE)} $\mathbf{R}^d_{\Theta}$ Due to the permutation invariance of the Transformer model and the scattered time $t$, we still need to integrate positional event information. To do so $\mathbf{Q, K}$ are rotated using the orthogonal matrix $\mathbf{R}^d_{\Theta}$ from \cite{Roformer} in function of the absolute event position $i$ in the sequence $S$. This method has two advantages: (1) it's not learnable (less likely to over-fitting), and (2) it integrates natively the relative position instead of altering $\mathbf{A}$ with a learnable bias like in \cite{relposatt}.
\end{itemize}
We make the distinction between our event-type embedding $\mathbf{E}$ and the other information per event (time and mileage) which we call context embedding $ \mathbf{CE} = \mathbf{T + M}$.




\subsubsection*{Continuous Time Mileage Aware Attention }\label{customattention}
We modify the vanilla Transformer from \cite{attention} by (1) adding the context embedding to the projected event-type embedding at every layer \cite{touvron2023llamaopenefficientfoundation}, and then (2) a Rotary Position Embedding (RoPE) \cite{Roformer} is applied to both query \(\mathbf{Q}\) and key \(\mathbf{K}\): 

\[
\mathbf{Q} = \mathbf{R}_{\Theta}^d (\mathbf{W}^Q \mathbf{E} + \mathbf{CE}),
\]
\[
\mathbf{K} = \mathbf{R}_{\Theta}^d (\mathbf{W}^K \mathbf{E} + \mathbf{CE})
\]

where \(\Theta = \{\theta_i = \theta_0^{-2(i-1)/d}, i \in [1, 2, \ldots, d/2]\}, \theta_0 = 10^4\).
More specifically, the inner product between query \(\mathbf{q}_m\) and key \(\mathbf{k}_n\) takes the event-type embedding \(\mathbf{e}_m, \mathbf{e}_n\) where \(m-n\) is their relative position with context \(\mathbf{CE}_m\) and \(\mathbf{CE}_n\):
{\small
\begin{align}\label{eq:qkproduct}
     \mathbf{q}^T_m \mathbf{k}_n &= (\mathbf{R}^d_{\Theta, m}(\mathbf{W}_q \mathbf{e}_m + \mathbf{CE}_m))^T 
    \mathbf{R}^d_{\Theta, n} (\mathbf{W}_k \mathbf{e}_n + \mathbf{CE}_n) \nonumber \\
    &= \mathbf{e}^T_m \mathbf{W}_q \mathbf{R}^d_{\Theta, n-m} \mathbf{W}_k \mathbf{e}_n + 
    \mathbf{e}^T_m \mathbf{W}_q \mathbf{R}^d_{\Theta, n-m} \mathbf{CE}_n + \nonumber \\
    &\quad \mathbf{CE}^T_m \mathbf{R}^d_{\Theta, n-m} \mathbf{W}_k \mathbf{e}_n + 
    \mathbf{CE}^T_m \mathbf{R}^d_{\Theta, n-m} \mathbf{CE}_n \nonumber \\
    &= \text{(1): query-to-key} + \text{(2): query-to-ce} \nonumber \\
    &\quad \text{(3): ce-to-key} + \text{(4): ce-to-ce}\nonumber \\
\end{align}}
where $\mathbf{R}^d_{\Theta, n-m} = (\mathbf{R}^d_{\Theta, m})^T\mathbf{R}^d_{\Theta, n}$ is a sparse orthogonal matrix. The additional terms (2), (3), (4) provide richer query and key representations when computing the attention scores.
Thus modifying Equation~\ref{eq:vanillattention}:

{\small
\[
\mathbf{A} = \text{softmax}\left( \frac{(\mathbf{R}_{\Theta}^d (\mathbf{W}^Q \mathbf{E}+ \mathbf{CE})) (\mathbf{R}_{\Theta}^d (\mathbf{W}^K \mathbf{E} + \mathbf{CE}))^T}{\sqrt{3d}}\right)
\]
}

Adding \(\mathbf{CE}\) after the projection to query and key can be seen as a refinement of 
\(\mathbf{Q}\), \(\mathbf{K}\) by \(\mathbf{T}\), \(\mathbf{M}\), providing additional context to the attention scores. We also add a scaling factor of 3 to compensate for the additional terms in Equation \ref{eq:qkproduct}.
\subsection{Multi-task Learning}
\begin{figure}[t]
    \centering
\includegraphics[width=0.86\columnwidth]{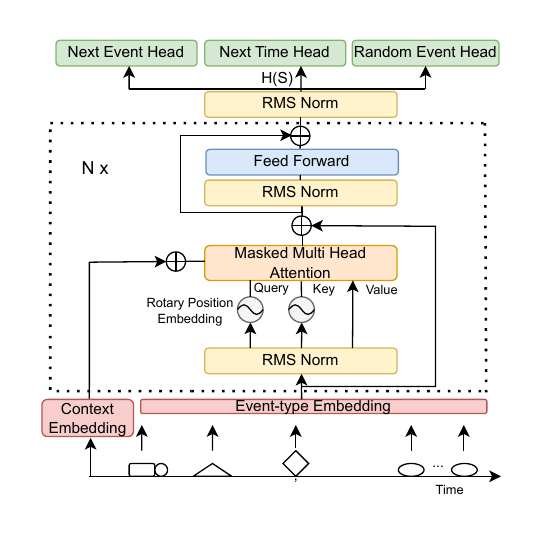}
    \caption{CarFormer architecture}
    \label{fig:carformer}
\end{figure}
\subsubsection{Next Event Prediction.} 
We use a standard language modelling objective which aims to minimize the cross-entropy loss between our output distribution $\hat{u}_i$ generated by our model's \textit{Next Event Head} and the next event $u_{i+1}$. \cite{shou2024selfsupervisedcontrastivepretrainingmultivariate} used a BERT \cite{bert} model trained on a masked event modelling task which is commonly used for bidirectional models. However, they simultaneously applied a causal mask resulting in a loss of the bidirectional property. We argue that by doing so, we lose a lot of sample efficiency, thus we will stick to a standard next token prediction.
$\hat{u}_i \in \mathbb{R}^{V}$ is the predicted probability distribution by the \textit{Next Event Head}, which integrates an RMS normalization \cite{rmsnorm} and one linear layer. 
The cross-entropy loss between $\hat{u}_i$ and $u_{i+1}$ (= a one-hot vector in $\{0,1\}^V$) is obtained by:
\begin{equation}
    \label{eq:nexteventtype}
    \mathcal{L}_c := -\sum_{i=0}^L \sum_{j=0}^V u_{i+1, j}  \log(\hat{u}_{i,j}) (1-\delta_{i,r})
\end{equation}
where $\delta_{i,r}$ is the Kronecker delta, which equals $1$ when $i = r, r \in R$ (set of randomly generated events) and $0$ otherwise. 

\subsubsection{Next Event Time Prediction.} In addition, we compute the Huber loss \cite{regloss} between the estimated inter-event time $\Delta \hat{t'}_i$ for the event $u_i$ and the ground truth $\Delta t'_i = f_t(t_{i+1}, 10) - f_t(t_i, 10)$ to deal with outliers and prevent exploding gradients with $\beta=1, \epsilon_i = \Delta \hat{t}'_i - \Delta t'_i$. 
\begin{equation}
\mathcal{L}_t := \sum^L_{i=0} (1-\delta_{i,r})\begin{cases} 
0.5 \epsilon^2_i  & \text{if } |\epsilon_i| < \beta, \\
 (|\epsilon_i| - 0.5) & \text{otherwise},
\end{cases}
\end{equation}
$t'$ is obtained using a log, hence we are essentially computing a kind of Mean Squared Logarithmic Error (MSLE) but with a $\beta$, useful to stabilize training and help convergence.

\subsubsection{Random Event Prediction.}\label{randomevent} Finally, a binary classifier that predicts whether an event was true or randomly generated is added. We motivate this choice by several papers \cite{shou2024selfsupervisedcontrastivepretrainingmultivariate, guofakeevent} stating that a model should learn when an event does not happen to reinforce the negative evidence of no observable events within each inter-event.
At each step \( i \), a random event is injected with probability \( p \). If a random event is successfully injected, the process continues until a failure occurs (i.e., the event is not injected), this allows for multiple random events to be injected in a row, allowing for more complexity. A detailed explanation of the injection process can be found in Algorithm \ref{alg:randomeventinjectionalgorithm} and Appendix \ref{appendix:pretraining}. The $\mathcal{L}_r$ loss is defined as the binary cross-entropy loss between the probability distribution $\hat{y}^r_i$ generated by our \textit{Random Event Head} and the ground truth $y^r_i$ at event step $i$:
\begin{equation}
\mathcal{L}_r := -\sum^{|R|}_{i=0} y^r_i \log(\hat{y}^r_i) + (1-y^r_i) \log (1-\hat{y}^r_i)
\end{equation}
\subsubsection{Total Loss.}
The total loss is defined as follows: 
\begin{equation}
 \mathcal{L} =  \frac{1}{L- |R|}(\mathcal{L}_c + \alpha \mathcal{L}_t) + \beta \frac{1}{|R|} \mathcal{L}_r
\end{equation}
where $\alpha, \beta$ are trade-off between the different loss, $L$ the sequence length, $R$ the set of random events injected in $S$.

\section{EPredictor}\label{epredictor}
Predicting only the next DTC in a\ sequence of DTC faults has its inherent limitation and remains a difficult task. \cite{Hafeez2024DTCTranGruIT} 
use DTCs those ECU, Base-DTC and Fault-Byte data have a cardinality of 83, 419 and 64, respectively, and only report a $81\%$ top-5 accuracy for next DTC prediction. This is because DTCs are not always correlated nor have causal links.
Instead, we are also using repair and warranty data 
to predict more important events such as EPs (error patterns). Repair and warranty data differ from DTCs since they are manually defined by domain experts after observing all DTCs and characterize a whole sequence $S$ and not an individual event $u_i$. Hence, we can define and say that a certain error pattern $y$ has happened at index $i=L$ in a sequence $S$. Note, that 
multiple EPs can occur at the same time. We can now define a supervised multi-label classification learning problem of predicting EPs.
With EPredictor, we leverage the \textit{seq2seq} nature of Transformers, where CarFormer outputs a sequence $H(S)$ of tokens encoded in a high-dimensional space $d$, positioning tokens with similar characteristics nearby. Then, this hidden representation $H(S)$ is fed into EPredictor which acts as an autoregressive multi-label classifier for EPs. We approach EP prediction as a machine translation task. By utilizing the contextualized hidden states $H(S)$ from CarFormer as key and value (i.e., through \textit{cross-attention}), this effectively transitions our model from a "seq2seq" to a "dtc2errorpattern" framework.
\begin{figure}[t]
    \centering
    \includegraphics[width=0.80\columnwidth]{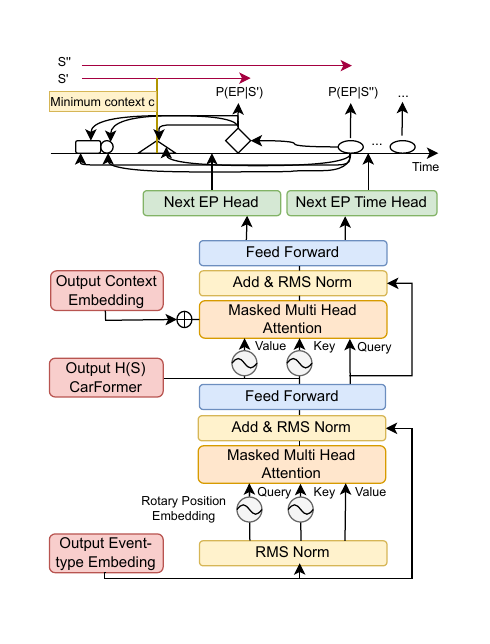}
    \caption{EPredictor architecture}
    \label{fig:epredictor}
\end{figure}

\subsection{Multi-Label Event Prediction}
Multi-label classification has recently gained interest in event prediction. \cite{multilabelpredictionfault} uses an LSTM for fault detection. More recently, \cite{Shou2023ConcurrentMP} considers concurrent event predictions as a multi-label classification and models such data with a Transformer architecture. 

To define the multi-label event prediction task with $N$ labels ($\equiv$EPs), we reuse each $S = \{(u_i, t_i ,m_i)\}^L_{i=0}$ and attach a binary vector $\mathbf{y} \in [0,1]^N$ to indicate the EPs occurring at time $t_L$. It's important to note that \(\mathbf{y}\) is invariant per sequence $S$, meaning for all events within $S$ the ground truth $\mathbf{y}$ will be the same. 
By using a causal mask on both CarFormer and EPredictor, we enable predictive maintenance predictions since tokens can only attend to the previous ones, as shown in Figure \ref{fig:epredictor}.
In our case, and in most real-world problems, EPs are highly imbalanced across our dataset, which is a considerably more challenging problem for our event prediction task, especially when the least occurring classes are the most important to detect \cite{multilabelpredictionfault}. In natural language processing, traditional up-sampling methods involve perturbation of $S$ by shuffling and replacement of tokens. However, in event data, we cannot afford to lose spatial and temporal information. Therefore, we inject random events $(u_i, m_i, t_i)$ with the same probability of $p=0.05$ as in Section \ref{sec:pretraining} and reuse the associated Algorithm \ref{alg:randomeventinjectionalgorithm}. We up-sample the different EPs classes up to a minimum $\theta_1 = 6000 $ and downsample the most popular ones down to a maximum $\theta_2 = 12000$. We drop also classes below 100 apparitions across the dataset.
We define a minimum context $c=30$  which acts as a “minimum history” of DTCs to retain. 
The \textit{Next EP Head} output a vector of probabilities  $\hat{y}_i = \text{sigmoid(MLP}_c( H_i))$ for each history $\{H_1, ..., H_i\}$, $i \in {c,...,L}$ where $H_i \in \mathbb{R}^d$ is the generated hidden representation from EPredictor at step $i$. For the regression task we forecast the time till the EP(s) occurrence \( \Delta \hat{t'}_i \) = $\text{MLP}_t$(\( H_i\)) where the ground truth is \( \Delta t'_i = f_t(t_L, 30) - f_t(t_i, 30)\).
Formally, we define our binary cross-entropy loss over the $N$ possible EPs for one step $i$ as follows:
\begin{equation}
\mathcal{L}^{ep}_i := -\frac{1}{N}\sum^{N}_{j=0} y_{j} \log(\hat{y}_{i,j}) + (1-y_{j}) \log (1-\hat{y}_{i,j})
\end{equation}
The total loss across $S$ with $L$ events and context $c$ is:
\begin{equation}
\mathcal{L}^{ep} := \frac{1}{L-c}\sum^{L}_{i=c} \mathcal{L}^{ep}_i 
\end{equation}

We use the Huber loss \cite{regloss} and define $\epsilon_i = \Delta \hat{t}'_i - \Delta t'_i$, with $\beta=1$ thus:
\begin{equation}
\mathcal{L}^t := \frac{1}{L-c} \sum^{L}_{i=c} \begin{cases} 
0.5 \epsilon^2_i  & \text{if } \epsilon_i < \beta, \\
|\epsilon_i| - 0.5 & \text{otherwise},
\end{cases}
\end{equation}
Our final loss to minimize is then : $\mathcal{L} = \mathcal{L}^{ep} + \gamma \mathcal{L}^ t$
\section{Experiments}\label{experiments}
We implemented and trained CarFormer and EPredictor models using PyTorch, the code is publicly available. Training details are included in Appendix \ref{appendix:pretraining},\ref{appendix:error_pattern}.

\subsection{Ablation I: CarFormer Embeddings}
Multiple CarFormer models with different choices of embeddings were evaluated on the next token prediction accuracy (ACC) and on the regression task with mean absolute percentage error (MAPE) and root mean square error (RMSE) to determine the best working CarFormer model. The different choices were  \textit{rot} (RoPE), \textit{time} (absolute time embedding added to the input \(\mathbf{U}\)), \textit{mileage} (also added to \(\mathbf{U}\)), \textit{m2c}, and \textit{c2m} (Appendix \ref{appendix:pretraining}).
In our case, we are more interested in the next event prediction task, thus we accept to lose some MAPE \% over the ACC (\%).
\begin{table}[ht]
  \centering
  \begin{tabular}{lccc}
    \toprule
    \textbf{Model}&\textbf{ACC(\%)}&\textbf{MAPE(\%)}&\textbf{RMSE}\\
    \midrule
    \textbf{rot-ce} & \textbf{22.64} & 3.2 & 0.04770 \\
    time & 21.48 & \textbf{2.9} & \textbf{0.04762} \\ 
    time-mileage & 21.38 & 3.0 & 0.04785 \\
    time-c2m-m2c & 21.58 & 3.5 & 0.04794 \\
    time-m2c & 21.52 & 3.6 & 0.04823 \\
    GPT & 19.89 & - & - \\
    \bottomrule
  \end{tabular}
  \caption{Overall prediction performance of CarFormer with different embeddings. Best results are in bold.}
  \label{tab:pretraining_exp}
\end{table}

Using only \textit{time} gave the best MAPE (2.9) but not the best ACC (21.48), suggesting that other features might improve the model predictions. For the mileage integration, our intuition was that doing an early summation of the two embeddings ($\mathbf{T}, \mathbf{M}$) seemed to denature the input $\mathbf{U}$ (ACC of \textit{time-mileage} $<$ ACC of \textit{time}) but the mileage of the vehicle could help differentiate between different DTCs. We tried to modify the attention dot products which seemed to help the ACC a bit (\textit{time-c2m-m2c}, \textit{time-m2c}) but increased the MAPE drastically. So we fused it with $\mathbf{CE}$ directly in $\mathbf{Q, K}$. Then, by applying a RoPE to the transformed input, the ACC increased while preserving the RMSE, leading to the best performing model, namely: $\textit{rot-ce}$.
\begin{table*}[ht]
  \centering
  \begin{tabular}{lcccccc}
    \toprule
    \textbf{Model} & \textbf{Micro F1 
 (\%)}  & \textbf{MAPE (\%) } & \textbf{MAE} & 
 $\textbf{CPMWAUC}_{f1} \uparrow$ & $\textbf{CPMWAUC}_{mae} \downarrow$ & \\
    \midrule
    rotcross-query-key-ce-1-2 & 82.69 & 32.45 & 0.0268 & 52.80 & 0.888 \\
    rotcross-query-key-ce-2 & 82.69 & \textbf{31.18} & 0.0254 & 56.61 & 0.882 & \\
    rotcross-query-ce-2 & 82.69 & \textbf{31.18} & 0.0254 & 53.00 & 0.884 & \\
    rotcross-key-value-ce-2 & \textbf{84.38} & 33.47 & 0.0263 & 65.06 & 0.904 \\
    \textbf{rotcross-key-value-scaled-ce-2} & \textbf{84.38} & 31.44 & \textbf{0.0252} & \textbf{67.63} & \textbf{0.874} \\
    rotnocross-ce-1-2 & 80.74 & 37.61 & 0.0275 & 42.95 & 0.927 & \\
    cross-speed & 83.53 & 34.73 & 0.0260 & 49.28 & 0.877 \\
    cross-mixffn & 83.41 & 33.37 & 0.0270 & 54.39 & 0.891 \\
    time-cross-query & 83.34 & 35.89 & 0.0275 & 45.71 & 0.896\\
    \bottomrule
  \end{tabular}
  \caption{EPredictor evaluation results with different model architecture on the test set (no up- nor down-sampling)}
    \label{tab:epredictor_overall}
\end{table*}

\subsection{EPredictor Experiments}
We used the micro-F1 score \cite{articlemultilabel} to assess the performance of the multi-label classification. To better understand and enhance our model's predictive maintenance capabilities, we introduce the concept of \textit{Confident Predictive Maintenance Window (CPMW)} which represents the interval within which our model can make reliable predictive maintenance predictions (similar to the "prediction window" described in \cite{interactivefeature}).
We quantify this with the \textit{CPMW Area Under Curve} (\textit{CPMWAUC}). The F1 score, MAE and MAPE have been calculated on average for all observations in Table \ref{tab:epredictor_overall}, and additionally for each history $H^i_t$ in Figure \ref{fig:epredictor} to understand how each model performs with different numbers of observations. To monitor the predictive maintenance capability of each model, the $\text{CPMWAUC}_{\text{f1}}$ and $\text{CPMWAUC}_{\text{mae}}$ were computed. The Appendix \ref{appendix:error_pattern} contains the models and metrics definition.
\subsection{Ablation II: EPredictor Architecture \& CPMW}
\begin{figure}[ht]
    \centering
    \includegraphics[width=1\columnwidth]{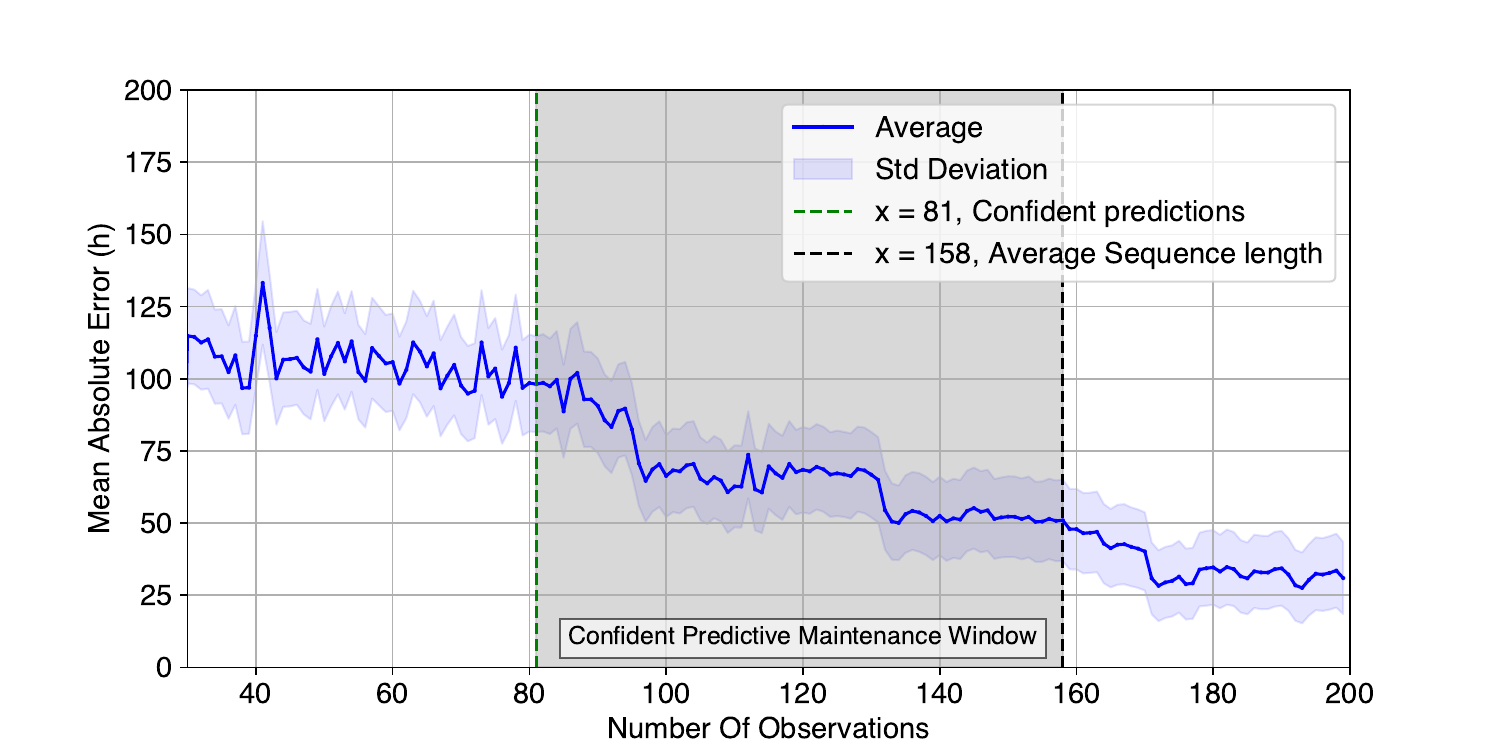}
    \caption{Evolution of the MAE in function of the number of observations for the best performing model.}
    \label{fig:mae_h}
\end{figure}

We explored several architectural changes and their impact on the CPMW: 
we applied a RoPE (\textit{rot}), a cross attention with \textit{query} or \textit{key} or \textit{value} to the second multi-head attention block (\textit{cross}), added the context embedding $\mathbf{CE}$ (\textit{ce}) to layer \textit{1} and/or \textit{2}, applied a scaling factor of $\sqrt{3d}$ to $\mathbf{Q, K}$ as shown in Equation \ref{eq:vanillattention} (\textit{scale}), injected a relative matrix $\mathbf{S}_{rel}$ into the attention scores (\textit{speed}),
and applied a mixed feed-forward network (\textit{mixffn}) \cite{NEURIPS2021_64f1f27b} to the mileage embedding. The \textit{time} model refers to $\mathbf{T}$ which is also added to $\mathbf{E}$. Finally, we trained a model with and without the \textit{Random Event Head}. 
Our experiments revealed several key insights:
By applying a \textit{cross} attention, we can see improvement in all metrics (\textit{rotnocross-ce-1-2}), which is consistent with the machine translation analogy \textit{"dtc2errorpattern"}. The best cross attention results were shown with $H(S)$ used as \textit{key-value}. Adding the mileage via an MLP layer (\textit{cross-mixffn}) seemed to help the MAPE (-2.5\%), the $\text{CPMWAUC}_{\text{f1}}(+9)$ and also the $\text{CPMWAUC}_\text{mae}(-0.005)$ compared to \textit{time-cross-query} model, suggesting that mileage is beneficial for both task. This makes sense since EPs are also dependent on the traveled distances between DTCs and the different stationary behavior of the vehicle. Furthermore, models incorporating a RoPE (\textit{rot}) performed significantly better in both regression and classification tasks like in the pre-training, highlighting the performance of RoPE in machine translation tasks \cite{Roformer}. Adding $\mathbf{CE}$ to the last layer (\textit{ce-2}) gave the best results, as opposed to adding it to both layers (\textit{ce-1-2}). Surprisingly, doing feature engineering on $\mathbf{T}, \mathbf{M}$ with a \textit{speed} matrix $\mathbf{S}_\text{rel}$ didn't help the metrics, which could indicate some missing modalities (e.g. mileage) during training, thus $\mathbf{CE}$ is more adapted for real-world scenarios. We monitored the need for our \textit{Random Event Head} in Figure \ref{fig:randomhead} (Appendix \ref{appendix:cpmwauc}) and noticed $+1.2\%$ in the F1 Score and $+10.1$ in the $\text{CPMW}_{f1}$.
When taking the best performing model (\textit{rotcross-key-value-scaled-ce-2}), the model entered the CPMW after 81 observations i.e. \textbf{half of the sequence}. Within this window, the model obtained an error of $\mathbf{\approx 65 \pm 14}$\textbf{h} when estimating the time of EP occurrence (Figure \ref{fig:mae_h}), highlighting the model's predictive capability within the CPMW. Otherwise, the average absolute error across all observations was approximately $58.4 \pm 13.2$h. 
By experimenting with these modifications, we aimed to identify the optimal architecture for predictive maintenance. The findings reveal that cross attention, context embedding in the second layer, and scaled attention significantly improve performance within the CPMW.

\begin{figure}[ht]
    \centering
    \includegraphics[width=1\columnwidth]{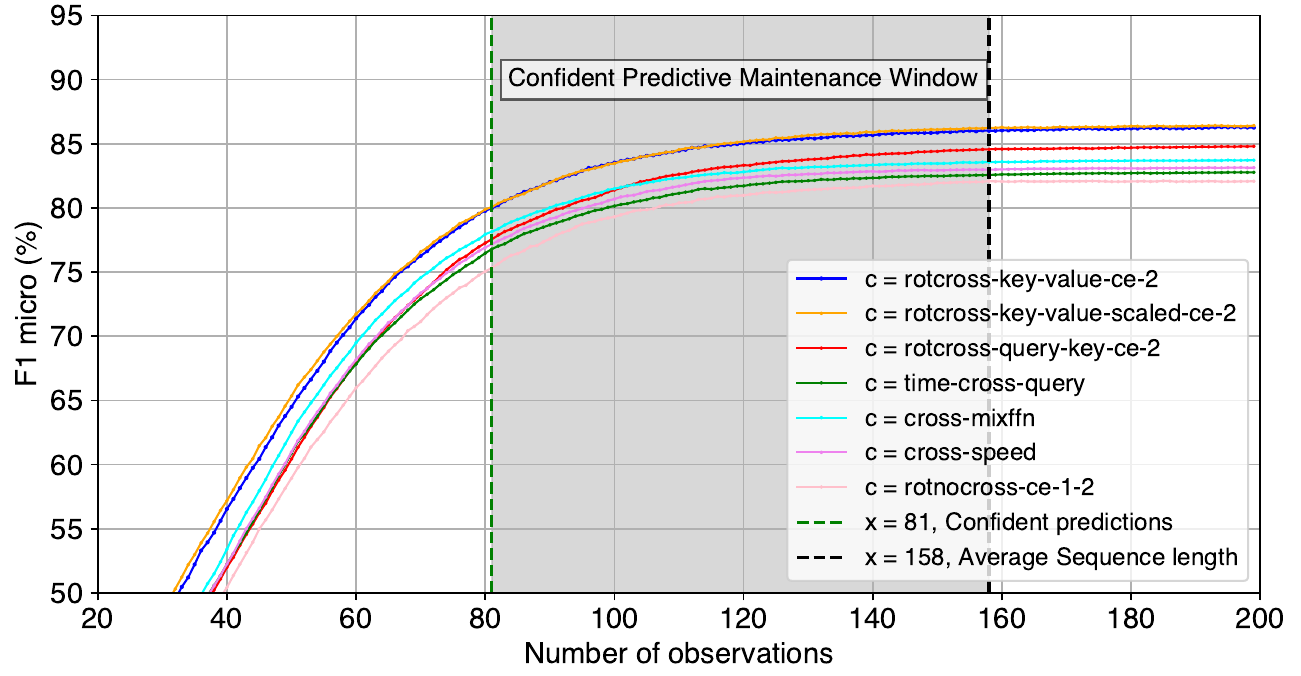}
    \caption{F1 Score comparison with multiple Epredictor architectures in function of the number of observations.}
    \label{fig:exp_epredictor_arch}
\end{figure}
\section{Conclusion}
This study bridges the gap between traditional event sequence modeling and fault event prediction in vehicles by using two language models. We have demonstrated through specific relevant metrics that our approach can accurately perform predictive maintenance by effectively predicting \textit{when} and \textit{what} error patterns are likely to occur, even with continuous, unbalanced, and high cardinality data.
In real-world settings, EPredictor is easy to use in a car. After each DTC occurrence and until reaching a minimum number of observations is reached, we would then infer the most likely EP and its time of occurrence. If the model is confident enough, the user will be alerted to an impending critical fault and directed to a nearby dealer, hence enhancing vehicle safety on the road and reducing maintenance costs.
\appendix
\section{CarFormer Pre-training Details}\label{appendix:pretraining}
We trained the CarFormer model for 70,000 steps with a learning rate of \(5 \times 10^{-4}\), scheduled using a cosine warm restart with 10,000 warm-up steps, and a weight decay of 0.1. The loss coefficients were set to \(\alpha = 1\) and \(\beta = 1\). The model architecture included 12 attention heads, 6 layers, and a feature size of 600, utilizing the GELU activation function in all feed-forward layers. We employed the AdamW optimizer with a batch size of 192 and a sequence length of 258, resulting in approximately 34 million parameters. It is important to note that on all experiments we fixed the same number of parameters when introducing news embeddings for all evaluated model to provide a fair comparison, same for Epredictor. The training data was split into $85\%$ training and $15\%$ testing without up-sampling, and random events were injected with a probability of \(p = 0.05\) per sample. On average, each training session lasted about 20 hours on an Nvidia A10G GPU.

\subsection{A.1 Models Definition}
A GPT \cite{gpt} model is added to improve comparison over baseline models.
We kept the original implementation of the feed-forward layers \cite{attention} but initialize the intermediate layers with: $ W_l \sim \mathcal{N}(0, \frac{2}{L \sqrt{d_l}}) $.
We used root-mean-square normalization (referred to as RMS Norm) from \cite{rmsnorm} instead of traditional layer normalization and initialize all other linear layers using the SMALLINIT schema \cite{transformernotears} $ W_l \sim \mathcal{N}\left(0, \sqrt{\frac{2}{d_l + 4d_l}}\right)$.
The names printed in the Tab \ref{tab:pretraining_exp} are model variations based on the embeddings given to CarFormer:
\begin{description}
  \item[time]: Along with the event type $\mathbf{E}$ an absolute time embedding is added such as input $\mathbf{U} = \mathbf{E} + \mathbf{T}$
  \item [mileage]: Same as time except we add a mileage embedding $\mathbf{M}$ such as $\mathbf{U} = \mathbf{E} + \mathbf{M}$
  \item [rot]: A RoPE \cite{Roformer} is applied to $\mathbf{Q,K}$ such as $\mathbf{Q} = \mathbf{R}^d_{\Theta}\mathbf{W}^Q\mathbf{E}$,  $\mathbf{K} =\mathbf{R}^d_{\Theta}\mathbf{W}^Q\mathbf{E}$
  \item [ce]: We add a context embedding $\mathbf{CE} = \mathbf{T} + \mathbf{M}$ to $\mathbf{Q,K}$ such as $\mathbf{Q = U W}^Q \mathbf{ + CE}$ to every attention layers.
  \item [m2c, c2m]: Inspired by the Disentangled Attention mechanism from DeBerta \cite{Deberta}. Attention score $\mathbf{A}_{i,j}$ between tokens $i$ and $j$ is computed from hidden states vector $\{\mathbf{H}_i\}$ at event step $i$ and a mileage vector  $\{\mathbf{M}_i\}$ at event step $i$ such as: $\mathbf{A}_{i,j} = \{\mathbf{H}_i, \mathbf{M}_{i} \} \times \{\mathbf{H}_j, \mathbf{M}_{j} \}^T = \mathbf{H}_i\mathbf{H}^T_j + \mathbf{H}_i\mathbf{M}^T_{j} + \mathbf{M}_{i}\mathbf{H}^T_j + \mathbf{M}_{i}\mathbf{M}^T_{j}$ = \textit{"content-to-content"} + \textit{"content-to-mileage"} + \textit{"mileage-to-content"} + \textit{"mileage-to-mileage"} = \textit{"c2c"} + \textit{"c2m"} + \textit{"m2c"} + \textit{"m2m"}. 
\end{description}
\subsection{A.2 Random Event Injection}
 Let \( A_i \) be the random variable representing the number of random events injected at step \( i \). We can say that: 
\begin{align*}
P(A_i = r) &= (1-p)p^r  \quad \text{for} \quad r = 0, 1, 2, \ldots
\end{align*}
Where \( A_i \) is the number of random events injected at step \( i \), \( r \) is the number of trials (injected events) until the first failure, \( p \) is the probability of injecting a random event. Hence, it's trivial to derive the expected number of random events injected per sequence \(S\) of length \(L\):
Therefore, the expected total number of injected fake events over the entire sequence of length \( L \) is:
$$
\mathbb{E}\left[\sum_{i=1}^{L} A_i\right] = Lp
$$ Which in our case would translate $\sim 150 \times 0.05 = 7.5$ random events per sequence.
\begin{algorithm}[t]
\caption{Random event injection algorithm}
\label{alg:randomeventinjectionalgorithm}
\textbf{Input}: Given a dataset $D$ with sequences S, each with length $L$, $S = \{(u_i, t_i, m_i)\}^L_{i=0}$ \\
\textbf{Parameter}: $p$ injection probability\\
\textbf{Output}: Dataset D'
\begin{algorithmic}[1] 
\STATE Let $D'=[]$.
\FOR{d $\leftarrow$ 1 \textbf{to} $D$}
\STATE S' = []
\STATE S = D[d]
\FOR{i $\leftarrow$ 1 \textbf{to} $L$ - 2}
    \STATE S'.append(S[i])
    \WHILE{$(p \leq \text{random.float}(0,1)) \And (i < (L - 2))$}
    \STATE $t'_i \sim \text{Unif}(t_i, t_{i+1})$
    \STATE $m'_i \sim \text{Unif}(m_i, m_{i+1})$
    \STATE $u'_i \sim \text{Unif}\{u_1, u_2, \ldots, u_n\}$
    \STATE S'.append($(u'_i, m'_i, t'_i)$)
    \ENDWHILE
    \IF{ $len(S') == (L - 2)$}
    \STATE \textbf{break}
    \ENDIF
\ENDFOR
   \STATE D'.append(S')
\ENDFOR
\STATE \textbf{return} D'
\end{algorithmic}
\end{algorithm}
\begin{figure}[t]
    \centering
    \includegraphics[width=1\columnwidth]{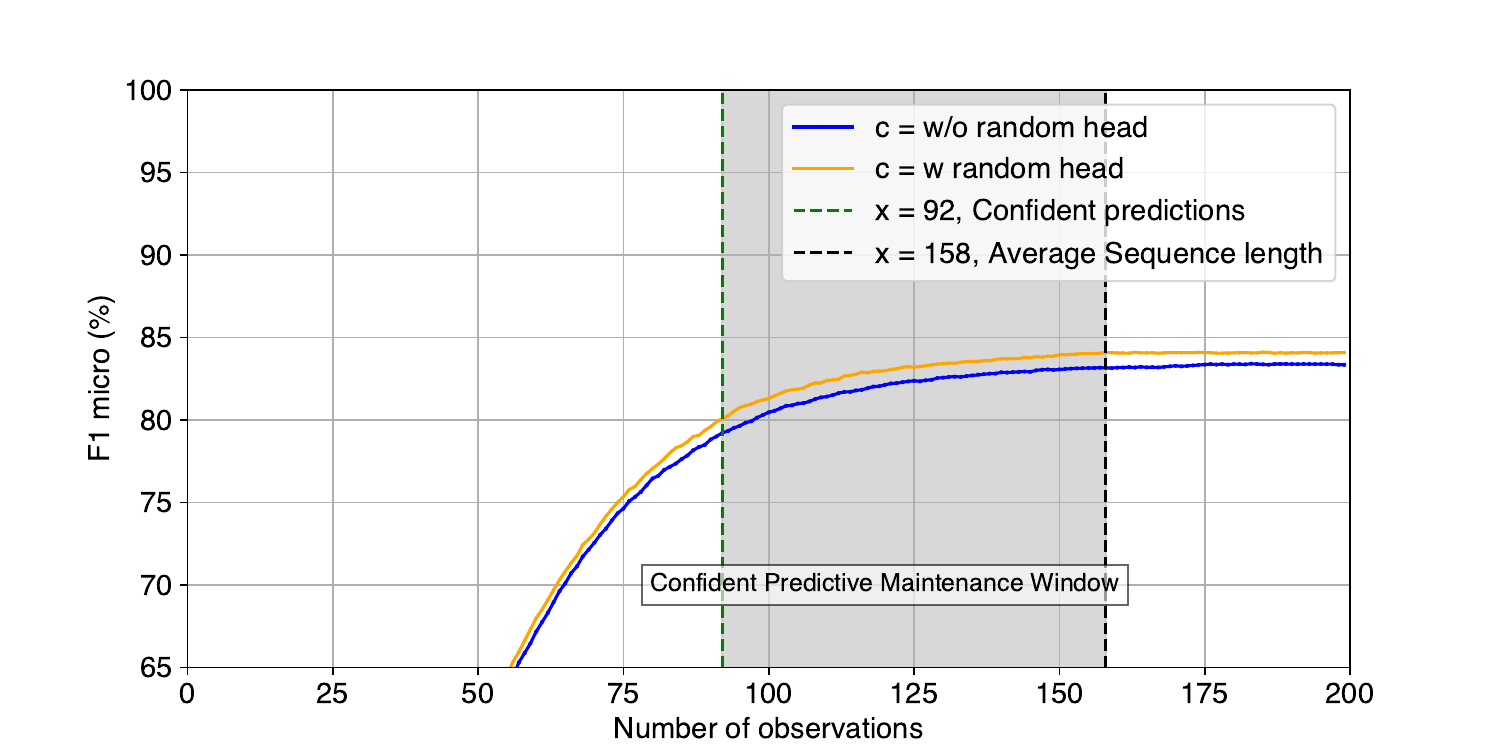}
    \caption{EPredictor CPMW with and w/o the random head.}
    \label{fig:randomhead}
\end{figure}
\section{EPredictor Training Details}\label{appendix:error_pattern}
We trained the EPredictor model for 10 epochs, using a patience of 2 on the evaluation loss to prevent overfitting. We used an AdamW optimizer with a learning rate of \(10^{-4}\), scheduled by a cosine warm restart. The model employed the GELU activation function in all feed-forward layers, with a batch size of 384 and \(\gamma = 1\) for the regression loss. The model architecture comprised attention layers with 12 attention heads each, and 2 feed-forward networks (FFN) with 1200 hidden neurons ($d_{model} \times 2$) each. The total number of trainable parameters (excluding the backbone) was 6.5 million per EPredictor. Each training session lasted about 6 hours on an Nvidia A10G GPU. We set \(c = 30\). The dataset was split into $70\%, 15\%$, and $15\%$ for training, validation, and testing, respectively.
The test set consists of 45,000 labeled sequences, with 269 unique error patterns (EP). A confidence threshold of 0.7 is used for classification.
For the regression task, when referring to MAE (Mean Absolute Error) or MAPE (Mean Absolute Percentage Error), we calculate the time $t'$ using the function $f_t(t, b)$. The MAE is defined as:
$$
\text{MAE} = \frac{1}{N}\sum_{i=1}^{L'} |\Delta \hat{t'}_i - \Delta t'_i|
$$
In Figure \ref{fig:mae_h}, the MAE (h) is calculated using the inverse function $f^{-1}_t: \mathbb{R} \to \mathbb{R}^+$, defined as $t = f^{-1}_t(t', b) = b^{t'+1} - 1 \forall t' \in \mathbb{R}$. Thus, the MAE (h) is:
$$
\text{MAE}_h = \frac{1}{N}\sum_{i=1}^{L'} |f^{-1}_t(\Delta \hat{t'}_i, 30) - f^{-1}_t(\Delta t'_i, 30)|
$$
For EPredictor, $L' = L - c$ and $i$ starts at $c$, whereas for CarFormer, $L' = L$ and $i = 0$.
\subsection{B.1 Architecture choice}
We defined multiple models for comparison with different architectures:

\begin{description}
    \item[mixffn]: Adds mileage to the first attention layer through a mix feed-forward network (MLP and 3x3 Conv) following \cite{NEURIPS2021_64f1f27b}. The formula is: $\mathbf{U'} = \mathbf{E} + \text{MLP(GELU(Conv}_{\text{3x3}}(\text{MLP(M)})$ where $\mathbf{E}$ is the output event embedding from CarFormer, and $\mathbf{U'} \in \mathbb{R}^{L \times d}$ is the input for the first attention layer of EPredictor.
    
    \item[rot]: Applies RoPE (Rotary Position Embeddings) to query ($\mathbf{Q}$) and key ($\mathbf{K}$) matrices as described in Section \ref{section:embeddings}.
    
    \item[speed]: Adds a relative speed matrix to the attention scores. Defines matrices $T_r$ and $M_r \in \mathbb{R}^L$ containing time and mileage features, forming $S = \frac{M}{T}$ and $S_{rel} = S - S^T \in \mathbb{R}^{L \times L}$. The attention scores are modified as $\mathbf{A} = \text{softmax}((\textbf{Q}\textbf{K}^T + \mathbf{S}_{rel})/\sqrt{2d})$. This model explores if incorporating relative speed information improves predictive performance.
\end{description}

\textit{Note: Some models such as rotcross-query-key-ce-1-2 are not shown in the figures to improve visual clarity}
\subsection{B.2 Context Size}
\subsubsection{Ablation III: Context Size}
\begin{figure}[t]
    \centering
    \includegraphics[width=1\columnwidth]{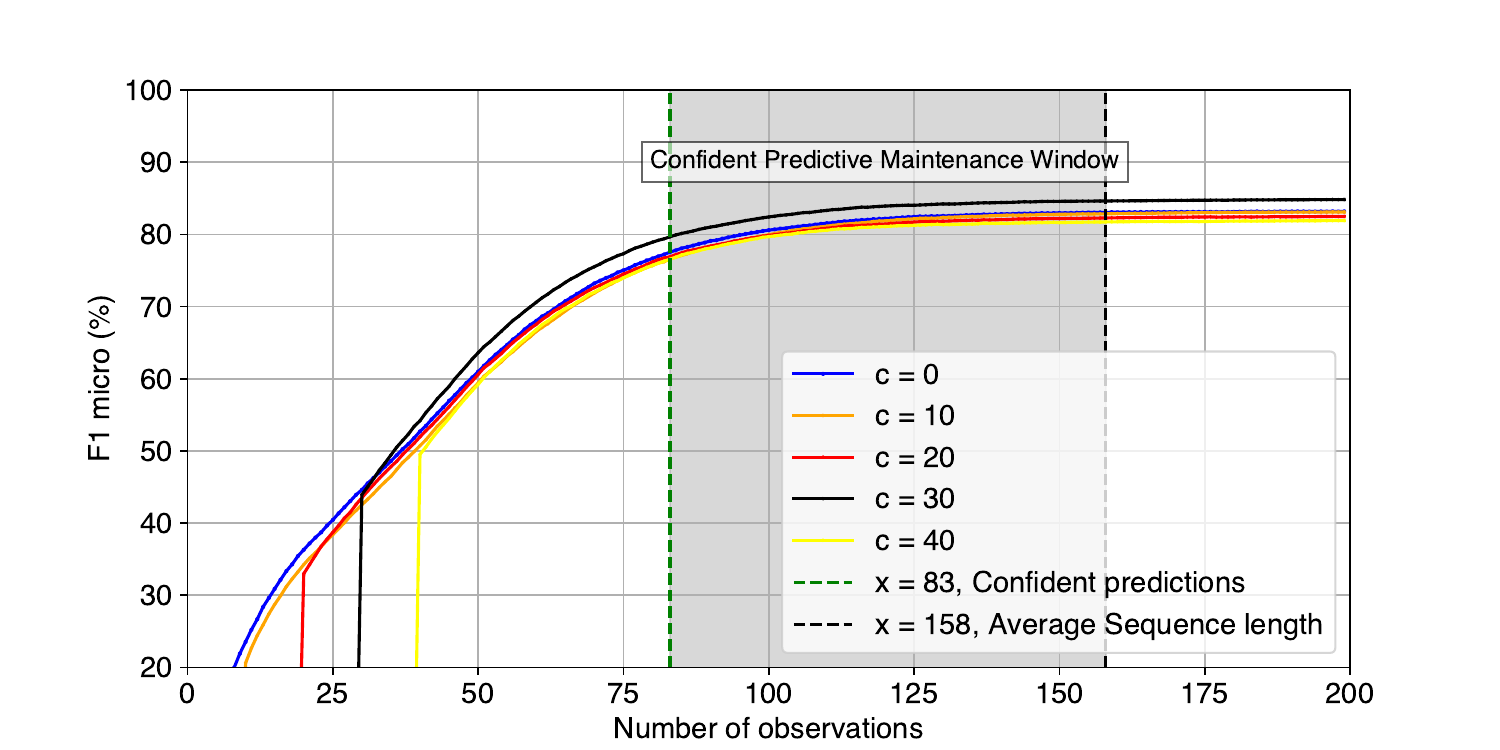}
    \caption{Evolution of the F1 Score with multiple $c$ in function of the number of observations.}
    \label{fig:exp_context}
\end{figure}

We explored the necessity of $c$ to provide EPredictor minimal observations before generating any predictions. Intuitively, with just 1 or 2 DTCs in $S$, it is not possible to predict future EPs. We experimented with several contexts $c = [0, 10, 20, 30, 40]$ on the test set.
Our experiments revealed $+2\%$ improvement in the F1 score using a context $c=30$. However, it is not trivial, as depicted in Figure \ref{fig:exp_context}, $c=10, 20, 40$ are worse than no $c$ at all.
We found that $c=30$ is beneficial for the $\text{CPMWAUC}_{f1}$ but should be used with caution and tested empirically. This indicates that increasing the context size might not always enhance the model's predictive performance and depends surely on the data distribution. 
\label{fig:aai_contextepredictor}
\subsection{B.3 Confident Predictive Maintenance Window}\label{appendix:cpmwauc}
The Confident Predictive Maintenance Window (CPMW) can be defined as the interval $[x_\theta, \mu_{seq} + \delta]$ where our predictive maintenance model is making prediction confidently on a set of sequences $S = \{(u_i, t_i, m_i)\}^L_{i=0}$.
\begin{itemize}
    \item $\theta$ is the threshold evaluation score (e.g., 80\% F1 score).
    \item $\mu_{\text{seq}}$ is the average sequence length.
    \item $\delta$ is the adjustment for noise (e.g., 8 for 5\% random events).
\end{itemize}
Confidently means at a typically chosen "high" or "low" score $\theta = \zeta(x_\theta)$ with $\zeta$ a metric function.
The CPMW Area Under the Curve or CPMWAUC of a given metric function $\zeta$ is given by its integral within the CPMW.
where $\zeta(x)$ continuous and integrable metric function.
\bibliography{aaai25}{}

\begin{thebibliography}{27}
\providecommand{\natexlab}[1]{#1}

\bibitem[{Bachmann and Nagarajan(2024)}]{bachmann2024pitfallsnexttokenpred}
Bachmann, G.; and Nagarajan, V. 2024.
\newblock The Pitfalls of Next-Token Prediction.
\newblock In \emph{ICLR 2024 Workshop: How Far Are We From AGI}.

\bibitem[{Brown et~al.(2020)Brown, Mann, Ryder, Subbiah, Kaplan, Dhariwal,
  Neelakantan, Shyam, Sastry, Askell, Agarwal, Herbert-Voss, Krueger, Henighan,
  Child, Ramesh, Ziegler, Wu, Winter, Hesse, Chen, Sigler, Litwin, Gray, Chess,
  Clark, Berner, McCandlish, Radford, Sutskever, and Amodei}]{gpt3}
Brown, T.; Mann, B.; Ryder, N.; Subbiah, M.; Kaplan, J.~D.; Dhariwal, P.;
  Neelakantan, A.; Shyam, P.; Sastry, G.; Askell, A.; Agarwal, S.;
  Herbert-Voss, A.; Krueger, G.; Henighan, T.; Child, R.; Ramesh, A.; Ziegler,
  D.; Wu, J.; Winter, C.; Hesse, C.; Chen, M.; Sigler, E.; Litwin, M.; Gray,
  S.; Chess, B.; Clark, J.; Berner, C.; McCandlish, S.; Radford, A.; Sutskever,
  I.; and Amodei, D. 2020.
\newblock Language Models are Few-Shot Learners.
\newblock In Larochelle, H.; Ranzato, M.; Hadsell, R.; Balcan, M.; and Lin, H.,
  eds., \emph{Advances in Neural Information Processing Systems}, volume~33,
  1877--1901. Curran Associates, Inc.

\bibitem[{Devlin et~al.(2019)Devlin, Chang, Lee, and Toutanova}]{bert}
Devlin, J.; Chang, M.-W.; Lee, K.; and Toutanova, K. 2019.
\newblock BERT: Pre-training of Deep Bidirectional Transformers for Language
  Understanding.
\newblock In \emph{North American Chapter of the Association for Computational
  Linguistics}.

\bibitem[{Du et~al.(2016)Du, Dai, Trivedi, Upadhyay, Gomez-Rodriguez, and
  Song}]{rrnembedding}
Du, N.; Dai, H.; Trivedi, R.; Upadhyay, U.; Gomez-Rodriguez, M.; and Song, L.
  2016.
\newblock Recurrent Marked Temporal Point Processes: Embedding Event History to
  Vector.
\newblock In \emph{Proceedings of the 22nd ACM SIGKDD International Conference
  on Knowledge Discovery and Data Mining}, KDD '16, 1555–1564. New York, NY,
  USA: Association for Computing Machinery.
\newblock ISBN 9781450342322.

\bibitem[{Gao et~al.(2020)Gao, Subramanian, Shanmugam, Bhattacharjya, and
  Mattei}]{guofakeevent}
Gao, T.; Subramanian, D.; Shanmugam, K.; Bhattacharjya, D.; and Mattei, N.
  2020.
\newblock A Multi-Channel Neural Graphical Event Model with Negative Evidence.
\newblock \emph{Proceedings of the AAAI Conference on Artificial Intelligence},
  34: 3946--3953.

\bibitem[{Hafeez, Alonso, and Riaz(2024)}]{Hafeez2024DTCTranGruIT}
Hafeez, A.~B.; Alonso, E.; and Riaz, A. 2024.
\newblock DTC-TranGru: Improving the performance of the next-DTC Prediction
  Model with Transformer and GRU.
\newblock \emph{Proceedings of the 39th ACM/SIGAPP Symposium on Applied
  Computing}.

\bibitem[{Hafeez, Alonso, and
  Ter-Sarkisov(2021)}]{faultpredmultivariatevehicule}
Hafeez, A.~B.; Alonso, E.; and Ter-Sarkisov, A. 2021.
\newblock Towards Sequential Multivariate Fault Prediction for Vehicular
  Predictive Maintenance.
\newblock In \emph{2021 20th IEEE International Conference on Machine Learning
  and Applications (ICMLA)}, 1016--1021.

\bibitem[{Hawkes(1971)}]{hawkeppp}
Hawkes, A.~G. 1971.
\newblock {Spectra of some self-exciting and mutually exciting point
  processes}.
\newblock \emph{Biometrika}, 58(1): 83--90.

\bibitem[{He et~al.(2021)He, Liu, Gao, and Chen}]{Deberta}
He, P.; Liu, X.; Gao, J.; and Chen, W. 2021.
\newblock DeBERTa: Decoding-enhanced BERT with Disentangled Attention.
\newblock In \emph{International Conference on Learning Representations}.

\bibitem[{Jadon, Patil, and Jadon(2024)}]{regloss}
Jadon, A.; Patil, A.; and Jadon, S. 2024.
\newblock A Comprehensive Survey of Regression-Based Loss Functions for Time
  Series Forecasting.
\newblock In Sharma, N.; Goje, A.~C.; Chakrabarti, A.; and Bruckstein, A.~M.,
  eds., \emph{Data Management, Analytics and Innovation}, 117--147. Singapore:
  Springer Nature Singapore.
\newblock ISBN 978-981-97-3245-6.

\bibitem[{Lin et~al.(2022)Lin, Wu, Zhao, Pai, and Li}]{lin2022exploring}
Lin, H.; Wu, L.; Zhao, G.; Pai, L.; and Li, S.~Z. 2022.
\newblock Exploring Generative Neural Temporal Point Process.
\newblock \emph{Transactions on Machine Learning Research}.

\bibitem[{Nguyen and Salazar(2019)}]{transformernotears}
Nguyen, T.~Q.; and Salazar, J. 2019.
\newblock Transformers without Tears: Improving the Normalization of
  Self-Attention.
\newblock In Niehues, J.; Cattoni, R.; St{\"u}ker, S.; Negri, M.; Turchi, M.;
  Ha, T.-L.; Salesky, E.; Sanabria, R.; Barrault, L.; Specia, L.; and Federico,
  M., eds., \emph{Proceedings of the 16th International Conference on Spoken
  Language Translation}. Hong Kong: Association for Computational Linguistics.

\bibitem[{Pirasteh et~al.(2019)Pirasteh, Nowaczyk, Pashami, L\"{o}wenadler,
  Thunberg, Ydreskog, and Berck}]{interactivefeature}
Pirasteh, P.; Nowaczyk, S.; Pashami, S.; L\"{o}wenadler, M.; Thunberg, K.;
  Ydreskog, H.; and Berck, P. 2019.
\newblock Interactive feature extraction for diagnostic trouble codes in
  predictive maintenance: A case study from automotive domain.
\newblock In \emph{Proceedings of the Workshop on Interactive Data Mining},
  WIDM'19. New York, NY, USA: Association for Computing Machinery.
\newblock ISBN 9781450362962.

\bibitem[{Radford and Narasimhan(2018)}]{gpt}
Radford, A.; and Narasimhan, K. 2018.
\newblock Improving Language Understanding by Generative Pre-Training.

\bibitem[{Shaw, Uszkoreit, and Vaswani(2018)}]{relposatt}
Shaw, P.; Uszkoreit, J.; and Vaswani, A. 2018.
\newblock Self-Attention with Relative Position Representations.
\newblock In Walker, M.; Ji, H.; and Stent, A., eds., \emph{Proceedings of the
  2018 Conference of the North {A}merican Chapter of the Association for
  Computational Linguistics: Human Language Technologies, Volume 2 (Short
  Papers)}, 464--468. New Orleans, Louisiana: Association for Computational
  Linguistics.

\bibitem[{Shchur et~al.(2021)Shchur, Türkmen, Januschowski, and
  Günnemann}]{nppreview}
Shchur, O.; Türkmen, A.~C.; Januschowski, T.; and Günnemann, S. 2021.
\newblock Neural Temporal Point Processes: A Review.
\newblock In Zhou, Z.-H., ed., \emph{Proceedings of the Thirtieth International
  Joint Conference on Artificial Intelligence, {IJCAI-21}}, 4585--4593.
  International Joint Conferences on Artificial Intelligence Organization.
\newblock Survey Track.

\bibitem[{Shou et~al.(2023)Shou, Gao, Subramanian, Bhattacharjya, and
  Bennett}]{Shou2023ConcurrentMP}
Shou, X.; Gao, T.; Subramanian, D.; Bhattacharjya, D.; and Bennett, K.~P. 2023.
\newblock Concurrent Multi-Label Prediction in Event Streams.
\newblock \emph{Proceedings of the AAAI Conference on Artificial Intelligence},
  37(8): 9820--9828.

\bibitem[{Shou et~al.(2024)Shou, Subramanian, Bhattacharjya, Gao, and
  Bennet}]{shou2024selfsupervisedcontrastivepretrainingmultivariate}
Shou, X.; Subramanian, D.; Bhattacharjya, D.; Gao, T.; and Bennet, K.~P. 2024.
\newblock Self-Supervised Contrastive Pre-Training for Multivariate Point
  Processes.
\newblock \emph{ArXiv}, abs/2402.00987.

\bibitem[{Su et~al.(2024)Su, Ahmed, Lu, Pan, Bo, and Liu}]{Roformer}
Su, J.; Ahmed, M.; Lu, Y.; Pan, S.; Bo, W.; and Liu, Y. 2024.
\newblock RoFormer: Enhanced transformer with Rotary Position Embedding.
\newblock \emph{Neurocomput.}, 568(C).

\bibitem[{Touvron et~al.(2023)Touvron, Lavril, Izacard, Martinet, Lachaux,
  Lacroix, Rozière, Goyal, Hambro, Azhar, Rodriguez, Joulin, Grave, and
  Lample}]{touvron2023llamaopenefficientfoundation}
Touvron, H.; Lavril, T.; Izacard, G.; Martinet, X.; Lachaux, M.-A.; Lacroix,
  T.; Rozière, B.; Goyal, N.; Hambro, E.; Azhar, F.; Rodriguez, A.; Joulin,
  A.; Grave, E.; and Lample, G. 2023.
\newblock LLaMA: Open and Efficient Foundation Language Models.
\newblock arXiv:2302.13971.

\bibitem[{Vaswani et~al.(2017)Vaswani, Shazeer, Parmar, Uszkoreit, Jones,
  Gomez, Kaiser, and Polosukhin}]{attention}
Vaswani, A.; Shazeer, N.; Parmar, N.; Uszkoreit, J.; Jones, L.; Gomez, A.~N.;
  Kaiser, L.~u.; and Polosukhin, I. 2017.
\newblock Attention is All you Need.
\newblock In Guyon, I.; Luxburg, U.~V.; Bengio, S.; Wallach, H.; Fergus, R.;
  Vishwanathan, S.; and Garnett, R., eds., \emph{Advances in Neural Information
  Processing Systems}, volume~30. Curran Associates, Inc.

\bibitem[{Xie et~al.(2021)Xie, Wang, Yu, Anandkumar, Alvarez, and
  Luo}]{NEURIPS2021_64f1f27b}
Xie, E.; Wang, W.; Yu, Z.; Anandkumar, A.; Alvarez, J.~M.; and Luo, P. 2021.
\newblock SegFormer: Simple and Efficient Design for Semantic Segmentation with
  Transformers.
\newblock In Ranzato, M.; Beygelzimer, A.; Dauphin, Y.; Liang, P.; and Vaughan,
  J.~W., eds., \emph{Advances in Neural Information Processing Systems},
  volume~34, 12077--12090. Curran Associates, Inc.

\bibitem[{Zhang and Sennrich(2019)}]{rmsnorm}
Zhang, B.; and Sennrich, R. 2019.
\newblock Root Mean Square Layer Normalization.
\newblock In Wallach, H.; Larochelle, H.; Beygelzimer, A.; d\textquotesingle
  Alch\'{e}-Buc, F.; Fox, E.; and Garnett, R., eds., \emph{Advances in Neural
  Information Processing Systems}, volume~32. Curran Associates, Inc.

\bibitem[{Zhang and Zhou(2014)}]{articlemultilabel}
Zhang, M.-L.; and Zhou, Z.-H. 2014.
\newblock A Review On Multi-Label Learning Algorithms.
\newblock \emph{Knowledge and Data Engineering, IEEE Transactions on}, 26:
  1819--1837.

\bibitem[{Zhang et~al.(2020{\natexlab{a}})Zhang, Lipani, Kirnap, and
  Yilmaz}]{selfatthawke}
Zhang, Q.; Lipani, A.; Kirnap, O.; and Yilmaz, E. 2020{\natexlab{a}}.
\newblock Self-Attentive {H}awkes Process.
\newblock In III, H.~D.; and Singh, A., eds., \emph{Proceedings of the 37th
  International Conference on Machine Learning}, volume 119 of
  \emph{Proceedings of Machine Learning Research}, 11183--11193. PMLR.

\bibitem[{Zhang et~al.(2020{\natexlab{b}})Zhang, Jha, Laftchiev, and
  Nikovski}]{multilabelpredictionfault}
Zhang, W.; Jha, D.~K.; Laftchiev, E.; and Nikovski, D. 2020{\natexlab{b}}.
\newblock Multi-label Prediction in Time Series Data using Deep Neural
  Networks.
\newblock \emph{CoRR}, abs/2001.10098.

\bibitem[{Zuo et~al.(2020)Zuo, Jiang, Li, Zhao, and
  Zha}]{transformerhawkeprocess}
Zuo, S.; Jiang, H.; Li, Z.; Zhao, T.; and Zha, H. 2020.
\newblock Transformer {H}awkes Process.
\newblock In III, H.~D.; and Singh, A., eds., \emph{Proceedings of the 37th
  International Conference on Machine Learning}, volume 119 of
  \emph{Proceedings of Machine Learning Research}, 11692--11702. PMLR.

\end{thebibliography}
\end{document}